# Active Visual Exploration Based on Attention-Map Entropy

**Adam Pardyl**[1,2,3] , **Grzegorz Rypeść**[1,4] , **Grzegorz Kurzejamski**[1] ,
**Bartosz Zieliński**[1,2,6] , **Tomasz Trzciński**[1,2,4,5]

[1]IDEAS NCBR
[2]Jagiellonian University, Faculty of Mathematics and Computer Science
[3]Jagiellonian University, Doctoral School of Exact and Natural Sciences
[4]Warsaw University of Technology
[5]Tooploox
[6]Ardigen

{adam.pardyl, grzegorz.rypesc, grzegorz.kurzejamski, bartosz.zielinski, tomasz.trzcinski}@ideas-ncbr.pl

## Abstract

Active visual exploration addresses the issue of limited sensor capabilities in real-world scenarios, where successive observations are actively chosen based on the environment. To tackle this problem, we introduce a new technique called Attention-Map Entropy (AME). It leverages the internal uncertainty of the transformer-based model to determine the most informative observations. In contrast to existing solutions, it does not require additional loss components, which simplifies the training. Through experiments, which also mimic retina-like sensors, we show that such simplified training significantly improves the performance of reconstruction, segmentation and classification on publicly available datasets.

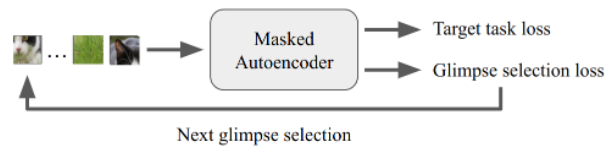
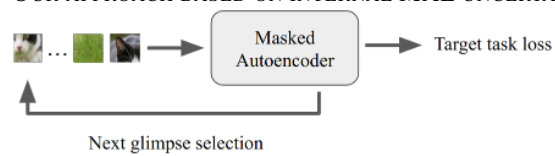

Figure 1: **Attention-Map Entropy (AME):** Our approach chooses the most informative observations by reusing the internal uncertainty coded in the attention maps. In contrast to existing methods, it does not require any auxiliary loss functions dedicated to active exploration. Therefore, the training concentrates on the target task loss, not on an auxiliary loss, which improves overall performance.

## 1 Introduction

Image-related tasks, such as image reconstruction, segmentation, or scene understanding, usually rely on the assumption of complete and unhindered access to input data [Girdhar *et al.*, 2022]. However, in many real-world applications, this assumption is not valid. For example, in the case of an embodied agent, such as a robot, its sensors have often limited registration capabilities, *e.g.* restricted field of view, while the environment is constantly changing [Wenzel *et al.*, 2021]. Additionally, the high computational cost of processing large amounts of data and the need for continuous movement further complicates the task of acquiring complete information about the environment. To reason about the whole environment, the agent needs to sample new observations in the most efficient way possible.[1]

Previous works in the field of embodied visual exploration have leveraged various sampling strategies to improve results on localization and mapping tasks [Ramakrishnan *et al.*, 2021]. Many of these methods use reinforcement learning [Ramakrishnan and Grauman, 2018] and multiple custom uncertainty measures [Ramakrishnan *et al.*, 2021], but they can be complex and difficult to train. To overcome this issue, [Seifi and Tuytelaars, 2019] introduced a streamlined architecture based on convolutional networks. They also simplified the setup for selecting image patches within a frame, referred to as glimpses, from 360° images in reconstruction tasks. The idea was further extended for segmentation task [Seifi and Tuytelaars, 2020]. In both scenarios, the choice of where to sample next was determined by using separate heads that output probability maps indicating the location of the next glimpse. In [Seifi *et al.*, 2021], they proposed a method of using attention pooling layers in a task-driven architecture to generate probability maps. However, this complex solution consisted of five separate training streams, each with its own optimization goals and a contrastive learning approach.

Furthermore, these existing methods involve separate training for the main task and the sampling decision strategy, which increases computational overhead of the entire system. [Jha *et al.*, 2023] addresses this limitation by using a transformer-based auto-encoder with a glimpse selection neural head. Nevertheless, all previous works rely on a dual approach to the active exploration problem. One target function is optimized to improve the quality of the main task, while another one is aimed at training the sampling decision strategy. This introduces unnecessary complexity of the model, increases the number of neural network components and yields more memory-heavy solutions.

---
[1]Supplementary material, including the source code, can be found at: https://github.com/apardyl/AME

In this paper, we address these complexity issues by introducing a transformer-based active vision method that leverages internal model uncertainty for next glimpse selection. Inspired by [Aloimonos *et al.*, 1988], we train our method to explore a scene by gathering partial observations about it sequentially without incorporating any separate sampling decision module. Contrary to the recent solution [Jha *et al.*, 2023], we use attention values trained inherently within the transformer-based decoder to generate decision-making priors, which provides superior performance over both random sampling and state-of-the-art solutions.

Our strategy for glimpse selection is based on the entropy of the transformer's attention maps and thus needs suitable architecture for the main task. Similarly to [Jha *et al.*, 2023], we base our network on masked autoencoder (MAE [He *et al.*, 2022]) to efficiently deal with the problem of partially available input.

We evaluate our approach using common benchmarks and relevant competing methods for reconstruction, classification, and segmentation tasks. Furthermore, we test our solution using both full-resolution patches, and retina-like glimpses [Sandini and Metta, 2003], publishing for the first time the empirical results for the latter data when coupled with transformer-based architectures.

We can summarize our contributions as follows:

- We present a new approach to glimpse selection in active visual exploration that leverages the internal transformer uncertainty stored in attention maps, eliminating the need for additional loss components.
- We validate our method with multiple patch configurations, including retina-like glimpses mimicking retina-like sensors.
- We thoroughly evaluate our solution on various visual tasks, demonstrating its superiority over the existing methods and various selection alternatives.

## 2 Related Works

**Missing data.** Our tasks are strictly connected with a missing data problem addressed by various research works. Some of them, like [Śmieja *et al.*, 2018], concentrate on fully connected networks where the missing data can be estimated with input data distribution. Other approaches concentrate on visual inpainting using adversarial and reconstruction loss [Pathak *et al.*, 2016]. [Li *et al.*, 2022] address the inpainting problem by using transformer architecture with a local masking scheme in attention layers, resulting in computation savings. [Naseer *et al.*, 2021] analyze well-known CNNs and claims they reach as low as 20% accuracy in ImageNet evaluation after masking only 20% of the image. Adequate progress in this field was made in Masked Auto-encoders (MAE) [He *et al.*, 2022] where authors use arbitrarily chosen masked parts of the input data to enrich visual features and regularize the training. Our method builds upon the MAE architecture, proven to achieve significant results in partially-masked image reconstruction.

**Active visual exploration.** Many works discuss simultaneous localization and mapping (SLAM) challenges in the context of active exploration [Ramakrishnan *et al.*, 2021]. [Rangrej and Clark, 2021] use the variational auto-encoder to hallucinate multiple hypotheses of the unseen parts of the image and infer uncertainty scores, which are further used to select the next glimpse parameters. Authors of [Seifi *et al.*, 2021; Seifi and Tuytelaars, 2020; Seifi and Tuytelaars, 2019] presented various solutions for active visual exploration, notably using CNN-based attention maps for glimpse selection and contrastive learning for global feature maps training. [Jha *et al.*, 2023] introduces MAE-based [He *et al.*, 2022] autoencoder with additional glimpse decision neural network to solve image reconstruction tasks. Results of these works have been reported in reconstruction, segmentation, and classification tasks, giving suitable baselines for our research. Our method solves the active visual exploration problem, but in contrast to the existing method, it uses auto-encoder internal, task-driven uncertainty to select successive glimpses.

**Glimpse Selection.** As opposed to solutions imposed by SLAM tasks, many researchers present glimpse selection problems in other contexts. For example, [Chai, 2019] adopts a deep reinforcement learning approach based on attention mechanisms to choose the best patch from consecutive video frames, yielding significant computational savings. [Jayaraman and Grauman, 2017] use uncertainty measures with reinforcement learning to estimate the following observations' parameters for a single 3D object analysis setup. [Rangrej *et al.*, 2022] introduces classification transformer architecture with an actor-critic module with an uncertainty estimator used for patch sampling. In our work, we actively search for successive glimpses defining the next patch sample using the entropy of the transformer attention maps. In opposition to [Rangrej *et al.*, 2022] or [Jha *et al.*, 2023], we do not need an additional neural network to make a glimpse decision. We also use no additional losses except those related to the target task.

## 3 Method

The fundamental idea of our approach is to leverage the internal uncertainty of the transformer-based autoencoder to select the most informative glimpses to be explored. Therefore, instead of using an additional loss function for glimpse selection, we propose a new method that obtains successive glimpses by analyzing the entropy of attention maps. This way, we can select the patch that has the highest entropy and confuses our model the most, which improves the convergence of our method and reduces its complexity compared to the existing methods.

As presented in Figure 2, our approach is based on a masked autoencoder with $T$ run-times. In the beginning, we run it for one glimpse, and at step $t < T$, there are $t$ known glimpses. To select the $t+1$ glimpse, we pass $t$ glimpses selected so far through the masked autoencoder and calculate entropy for its attention maps. The masked patch with the highest entropy is chosen as the next glimpse. The network produces the final output for the target task (*e.g.* a reconstructed image for a reconstruction task) for $t = T$. This selection process is described in Section 3.2. Moreover, in Section 3.1 and Section 3.3, we describe the transformer-based autoencoder and the adjustments of its architecture to various target tasks.

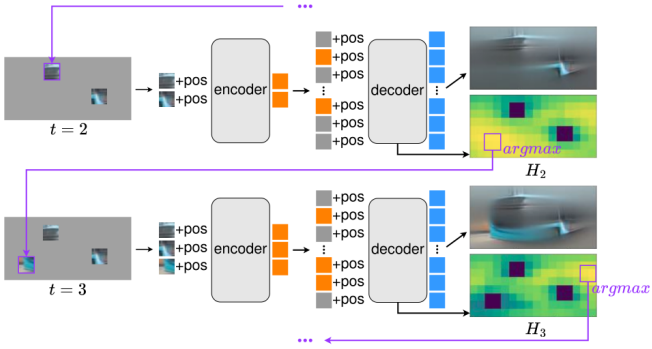
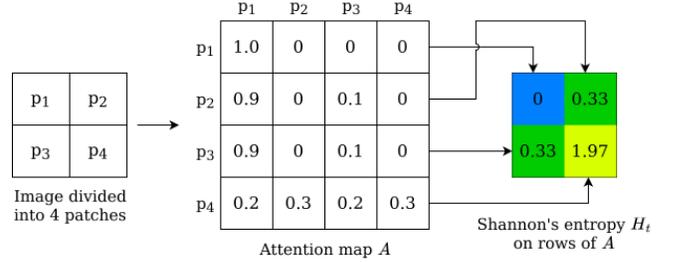

Figure 2: **Architecture for reconstruction:** The agent observed two patches of the image, which are processed by the encoder to produce their feature representations (orange rectangles). These outputs are combined with the masked patches (shown as gray rectangles) and passed through the decoder. The decoder reconstructs the missing image patches. Additionally, our method generates the entropy map for one of the decoder's multi-head self-attention layers and uses it to select the location of the third glimpse. The process repeats till we reach the assumed number of glimpses.

Figure 3: **Entropy map:** To explain the idea of the entropy map based on attention in the transformer layer, let us consider an image divided into four patches ($2 \times 2$) on the left. Its attention map will be a $4 \times 4$ matrix, where each row represents the attention weights used to calculate the output in the next transformer layer for a corresponding patch. Calculating Shannon's entropy for each row will result in a $2 \times 2$ entropy map. The patch with the highest entropy value is selected as the next glimpse.

### 3.1 Transformer-based Masked Autoencoder

Our approach employs a transformer-based Masked Autoencoder (MAE) network, as outlined in [He *et al.*, 2022], which includes an encoder and a decoder.

The encoder is a modified version of the Vision Transformer (ViT) [Dosovitskiy *et al.*, 2020]. As an input at a step $t$, it obtains a sequence of patches $p_1, p_2, ..., p_t : \forall_i p_i \in \mathbb{R}^{P^2 \times C}$ observed so far, where $P$ is the patch size, and $C$ is the number of input image channels. The patches are projected into $E$-dimensional embeddings using a linear layer and summed with positional embeddings. Then, they are processed through transformer blocks together with a learnable CLS token, resulting in $t$ latent embeddings of dimension $D$. Because the number of visible patches may vary between images in a batch, we employ transformer blocks that allow pad tokens similar to those proposed in [Vaswani *et al.*, 2017].

The decoder is a transformer with a linear head that operates on $t$ latent embeddings generated by the encoder and the coded positions of the $\frac{HW}{P^2} - t$ remaining patches, where $H$ and $W$ represent the height and width of the input image. The concatenated embeddings of dimension $D$ are positionally encoded and processed through transformer blocks. The last part of the decoder is a linear layer with dimensions varies depending on the task (see Section 3.3).

### 3.2 Glimpse Selection

To explore the internal uncertainty of transformer-based models, we analyze the entropy of the attention map [Vaswani *et al.*, 2017], as presented in Figure 3. For this purpose, we start with calculating attention map $A \in \mathbb{R}^{t \times t}$ for patches $p_1, p_2, ..., p_t$ obtained so far as

$$A = \text{softmax}(KK^T / \sqrt{d_k}), \quad (1)$$

where $K \in \mathbb{R}^{t \times D}$ is a matrix that packs together all embeddings of patches from one of the decoder's layers. The successive rows of $A$ contain weights used to calculate a weighted average of the patches outputs. Therefore, to obtain the output for patch $i$, we sum the values of all patches weighted with $A[i]$ ($i$-th row of $A$).

However, we can observe that the entropy of $A$ rows significantly differs, and it can be used to select an adequate candidate for the next glimpse. To explain how, let us consider patches 2 and 4 from Figure 3 with $A[2] = [0.9, 0, 0.1, 0]$ and $A[4] = [0.2, 0.3, 0.3, 0.2]$. While the 2nd patch output is obtained based mostly on the value of the 1st patch, the 4th patch output is obtained based on all other patches. Hence, the model is confident about the 2nd patch but is uncertain about patch 4. Therefore, the latter is a better candidate for the next glimpse, as it needs to be explored.

To formalize this idea, let us also recall that transformers use multi-head attention, which computes multiple attention maps, each for embeddings obtained with different linear projections. Therefore, assuming that $A_{h,t}$ denotes the attention map of $h$th head in step $t$, the entropy map for a given multi-head self-attention layer is calculated as

$$H_t = \sum_h H(A_{h,t}[i]), \quad (2)$$

where $H$ is the Shannon entropy. Then, we replace entropy values for known patches to 0 to omit them in the further selection process, and we choose the patch with the maximal value of $H_t$ as the next glimpse

$$i^* = \text{argmax}_{i=1,...,d_{model}} H_t[i]. \quad (3)$$

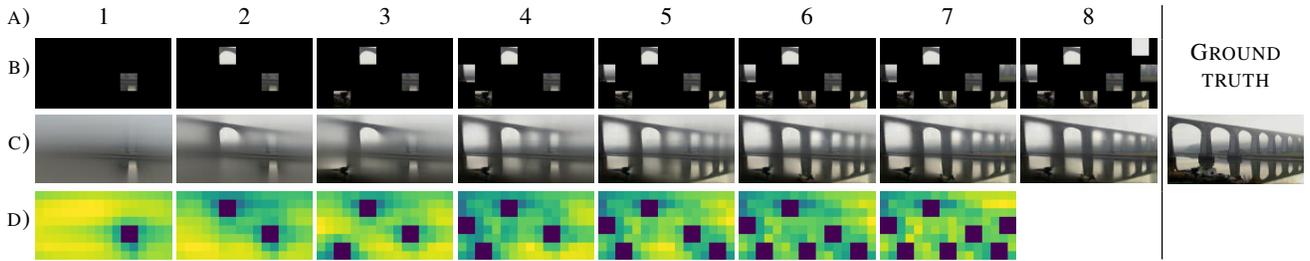

Figure 4: **Glimpse-based reconstruction step-by-step:** The figure shows a glimpse selection process based on AME for $8 \times 32^2$ glimpses for a sample $256 \times 128$ image. The rows correspond to A) step number, B) model input (glimpses), C) model prediction given, D) decoder attention entropy (known areas are explicitly set to zero). The algorithm explores the image in places where the reconstruction result is blurry.

### 3.3 Differences Caused by Final Task

For the reconstruction and segmentation tasks, the last layer of the decoder returns a representation of dimension $C' \times P^2$ for each processed patch, where $C'$ is the number of input image channels for reconstruction and the number of segmented classes for segmentation. In this case, the output for the CLS token is omitted. For the classification task, we add an auxiliary classification head, which takes the CLS token from the encoder as input to classify the image.

## 4 Experimental Setup

### 4.1 General considerations

**Architecture.** In all our experiments, we use 24 transformer blocks in the encoder, with an embedding size of 1024 (the same parameters as the ViT-L architecture [Dosovitskiy *et al.*, 2020]), and 8 decoder blocks with an embedding size of 512. We use the AdamW optimization algorithm [Loshchilov and Hutter, 2018] with a weight decay value of 0 for reconstruction and $10^{-4}$ for other tasks. The learning rate is first linearly brought up to $10^{-4}$ for the first 10 training epochs, then decayed with the half-cycle cosine rate to $10^{-8}$ for the rest of the training. We train the model for 75 epochs with early stopping. We augment the training data with random scaling, cropping, and horizontal flipping. Then we resize the image to the required size. Unless otherwise specified, we initialize the model with MAE [He *et al.*, 2022] weights trained on the ImageNet-1K [Deng *et al.*, 2009] dataset. For $256 \times 128$ images, we also fine-tune those MAE weights on the target resolution using the MS COCO dataset [Lin *et al.*, 2014] using the original MAE implementation. For glimpse selection, we extract the attention map from the last decoder block. The first glimpse is selected by the entropy mechanism run with mask tokens only (zero known patches).

**Datasets.** We evaluate our method on 3 different vision datasets. The largest one, MS COCO dataset (Common Objects in Context; 2014 split version) [Lin *et al.*, 2014], consists of 83K train images and 41K validation images. The second one, ADE20K [Zhou *et al.*, 2017] dataset, consists of 26K training and 2k validation images, containing 150 classes for the semantic segmentation task. The smallest one is SUN360 [Song *et al.*, 2015] dataset with approximately 8K 360° panoramic images, unevenly split between 26 classes for the multi-class classification task. As the last dataset does not have a predetermined train-test split, we use a 9 : 1 train-test split based on an index provided by authors of [Seifi *et al.*, 2021].

**Glimpse regimes and baselines.** We perform the experiments in two different glimpse regimes associated with two sets of state-of-the-art baselines.

First, we compare our method against fully convolutional approaches: Attend and segment [Seifi and Tuytelaars, 2020] and Glimpse, attend and explore [Seifi *et al.*, 2021]. Both methods use three-level retina-like glimpses, simulating the use of retina-like sensors introduced in [Sandini and Metta, 2003]. In this setup, an image is resized to $256 \times 128$ pixels, and the algorithm extracts 8 retinal glimpses of size 48x48 pixels. This corresponds to extracting 18.75% of pixels (accounting for retinal down-scaling) and 56.26% of the image area explored. As glimpses consist of multiple ViT patches, during the glimpse selection process, we average the glimpse selection map from Equation 2 over the size of a glimpse.

Secondly, we compare it to a transformer-based method called SimGlim [Jha *et al.*, 2023]. In this scenario, an image is resized to $224 \times 224$ pixels, and we extract 37 regular (non-retinal) glimpses of size 16x16 pixels (one ViT patch [Dosovitskiy *et al.*, 2020]). As a result, 18.75% of pixels are extracted, covering 18.75% of the image area.

The results of baseline methods (including visualizations) are copied from the original papers because we could not reproduce them due to incomplete or missing code.

We also conducted ablation experiments using two naive sampling methods. The first method is random sampling, which has the potential for overlap of glimpses. The second method samples glimpses from a fixed, non-overlapping checkerboard pattern [He *et al.*, 2022].

### 4.2 Tasks

**Reconstruction.** The reconstruction task involves recreating the ground-truth image with a limited number of glimpses. We evaluate our method on the MS COCO, ADE20K, and SUN360 datasets. The reconstruction quality is measured using the Root Mean Squared Error (RMSE), also used as a loss function.

**Classification.** We evaluate our model in the multi-class classification task on the SUN360 dataset using a limited number of glimpses. We use accuracy as a metric and cross entropy as a loss function. We evaluate our model in two training regimes: a) training the entire model for classification (*train-all*) and b) training only the classification head-on features generated by a model trained on the reconstruction task (*head-only*). The classification head consists of linear layers: one in the *train-all* scenario or two separated by a GELU activation layer for the *head-only* scenario. For the *train-all* scenario, we do a weighted sum of the classification loss (cross-entropy) and a loss of the decoder (required for the glimpse selection mechanism).

| METHOD | SUN360 | ADE20K | MS COCO | IMAGE RES. | GLIMPSE REGIME | PIXEL % | AREA % |
|---|---|---|---|---|---|---|---|
| ATTSEG | 37.6 | 36.6 | 41.8 | $128 \times 256$ | $8 \times 48^2$ (RETINAL) | 18.75 | 56.25 |
| GLATEX | 33.8 | 41.9 | 40.3 | $128 \times 256$ | $8 \times 48^2$ (RETINAL) | 18.75 | 56.25 |
| OURS (RETINAL) | **23.6** | **23.8** | **25.2** | $128 \times 256$ | $8 \times 48^2$ (RETINAL) | 18.75 | 56.25 |
| OURS (NON-RETINAL) | 37.9 | 40.7 | 43.2 | $128 \times 256$ | $8 \times 16^2$ (NON RET.) | 6.25 | 6.25 |
| OURS (NON-RETINAL) | 29.8 | 30.8 | 32.5 | $128 \times 256$ | $8 \times 32^2$ (NON RET.) | 25.00 | 25.00 |
| OURS (NON-RETINAL) | **20,1** | **20.6** | **22.1** | $128 \times 256$ | $8 \times 48^2$ (NON RET.) | 56.25 | 56.25 |
| SIMGLIM (DETACHED) | 26.2 | 27.2 | 29.8 | $224 \times 224$ | $37 \times 16^2$ (NON RET.) | 18.75 | 18.75 |
| SIMGLIM (END-TO-END) | 28.0 | 28.8 | 31.3 | $224 \times 224$ | $37 \times 16^2$ (NON RET.) | 18.75 | 18.75 |
| OURS (NON-RETINAL) | **23.4** | **26.2** | **28.6** | $224 \times 224$ | $37 \times 16^2$ (NON RET.) | 18.75 | 18.75 |

Table 1: **Reconstruction results:** Comparison of our model in reconstruction task against AttSeg [Seifi and Tuytelaars, 2020], GlAtEx [Seifi *et al.*, 2021] and SimGlim [Jha *et al.*, 2023] on SUN360, ADE20K and MS COCO datasets. The metric used is a root mean square error (RMSE; lower is better). For each experiment, we provide a training and evaluation regime defined by a number of glimpses of a specific resolution. Pixel % and area % denote respectively: the percentage of image pixels known to the model and the percentage of image area seen by the model. Differences in both measures occur when dealing with retina-like glimpses, which have lower pixel counts by design. Our method outperforms competitive solutions in all configurations.

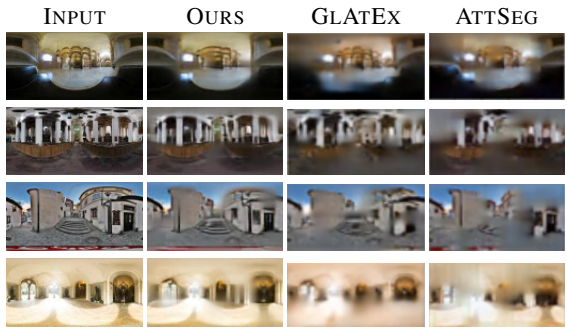

Figure 5: **Reconstruction quality for SUN360:** Reconstruction results of our method compared with AttSeg [Seifi and Tuytelaars, 2020] and GlAtEx [Seifi *et al.*, 2021] on the SUN360 dataset. Reconstructions done with our method are visibly better than the competition.

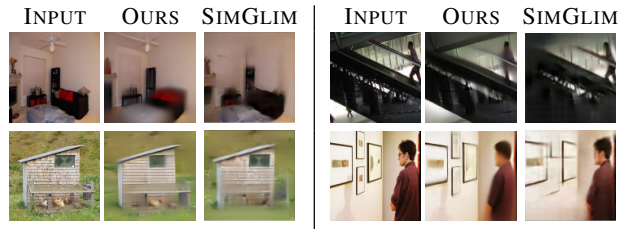

Figure 6: **Reconstruction quality for ADE20K:** Figure shows difference in reconstruction quality against SimGlim [Jha *et al.*, 2023] on the ADE20K dataset. SimGlim reconstructs a single object slightly better, but our method recovers more objects in the scene.

**Segmentation.** The goal of the segmentation task is to classify each pixel of the input image. Our method is tested on the ADE20K dataset. The mean Pixel Accuracy (mPA; averaged over classes), Pixel Accuracy (PA; averaged over all pixels), and Intersection over Union (IoU) are used as a metric, and the pixel-wise cross-entropy is used as a loss function. We initialize the transformer weights in two ways: using MAE or SETR (trained on ADE20k) weights as the source of encoder weights. In both cases, the decoder is initialized randomly.

## 5 Results

To illustrate how the glimpse selection process works step by step, we present it in Figure 4 on a sample image from the ADE20K dataset. As can be seen, with each glimpse, not only the glimpse neighborhood but the entire output image becomes clearer because the transformer gains a better understanding of the scene.

### 5.1 Reconstruction

In our study, we demonstrate superior performance in the retina-like glimpse setup for all datasets examined. Supporting results are presented in the top part of Table 1. Additionally, as shown in Figure 5, the reconstructions produced by our method are of higher quality, with visibly increased clarity and sharpness.

We also conduct tests without retina degradation of patches and show that using non-retinal glimpses improved results by over 10% of all scores. Moreover, our tests with smaller non-retinal patch sizes show that we still achieved better RMSE scores using less than half the area. In this experiment, we use eight patches in a 32x32 resolution, using 25% of the area compared to 56.25% in baseline, as seen in the middle part of Table 1.

When comparing our method to the transformer-based SimGlim [Jha *et al.*, 2023], we find that our solution further improves reconstruction scores across all of the databases.

| METHOD | ACCURACY (%) |
|---|---|
| ATTSEG | 52.6 |
| GLATEX (FULL) | 56.4 |
| GLATEX (NO DECODER) | 67.2 |
| OURS (HEAD-ONLY) | 70.1 |
| OURS (TRAIN-ALL) | **75.7** |

Table 2: **Classification results:** Comparison of our model's classification performance against AttSeg [Seifi and Tuytelaars, 2020] and GlAtEx [Seifi *et al.*, 2021] on the SUN360 dataset. The metric used is accuracy (higher is better). Our method outperforms competitive methods in both Train-all and Head-only training options.

| METHOD | MPA(%) | PA(%) | IoU(%) |
|---|---|---|---|
| ATTSEG | - | 47.9 | - |
| GLATEX | - | 52.4 | - |
| OURS (MAE-WEIGHTS) | 32.2 | **70.27** | 24.4 |
| OURS (SETR-WEIGHTS) | **35.6** | 69.5 | **27.6** |

Table 3: **Segmentation results:** Comparison of our model against AttSeg [Seifi and Tuytelaars, 2020] and GlAtEx [Seifi *et al.*, 2021]. The metric used is mean Pixel-wise Accuracy (mPA, higher is better), Pixel-Accuracy (PA, higher is better), and Intersection over Union (IoU, higher is better). Our solution outperforms competitive methods on all metrics.

This is shown in the bottom part of Table 1. It is noteworthy that the authors of [Jha *et al.*, 2023] also used MAE [He *et al.*, 2022] architecture for the encoder and decoder. Therefore, improvement is mostly achieved from the glimpse selection method we propose rather than the underlying architecture.

In the qualitative comparison shown in Figure 6, we can observe that while SimGlim produces more detailed reconstructions of individual objects, our method can discover and reconstruct more objects in the image overall, indicating better exploration capabilities. Moreover, authors of [Jha *et al.*, 2023] analyzed attention-based solutions for glimpse selection and concluded that it was not performing better than random sampling. However, in our study, we observe that our more sophisticated way of using this information outperforms random approaches. We provide further discussion on these ablation studies in Section 5.5.

### 5.2 Classification

Our method showed improved performance compared to the solutions presented in [Seifi *et al.*, 2021] and [Seifi and Tuytelaars, 2020] in both training modes (train-all and head-only), as presented in Table 2. These results align with the improvements observed in the reconstruction task. The use of transformer-based architecture in the underlying MAE [He *et al.*, 2022] is likely a significant contributing factor to these improvements results, as we achieved better results even with auto-encoders weights frozen during training (head-only).

### 5.3 Segmentation

Consistently, our method obtains better results in the segmentation task than compared, state of the art methods. As presented in Table 3, our approach achieves higher PA scores than those reported in [Seifi and Tuytelaars, 2020] and [Seifi

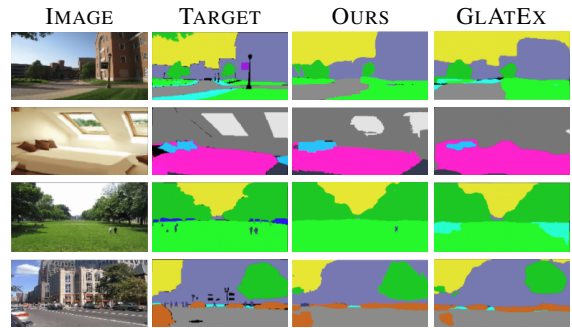

Figure 7: **Segmentation quality for ADE20K:** Semantic segmentation results of our method compared with GlAtEx [Seifi *et al.*, 2021] on the ADE20k dataset. Qualitatively, segmentation maps produced by our method are at least as good as those of the competition.

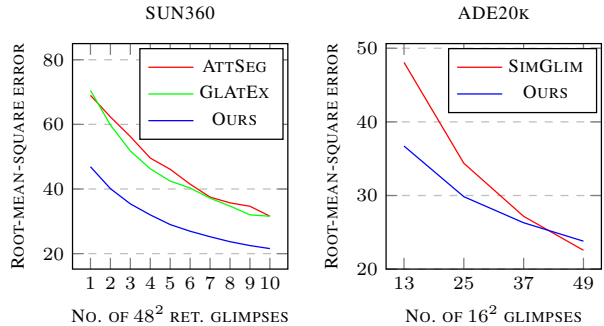

Figure 8: **Comparative reconstruction results in regards to glimpse selection step**: Figures present the effect of the number of glimpses on the reconstruction task. On the left, we compare our model against AttSeg [Seifi and Tuytelaars, 2020] and GlAtEx [Seifi *et al.*, 2021] on the SUN360 dataset in $48^2$ retinal glimpse scenario. On the right, we compare our model against SimGlim [Jha *et al.*, 2023] on ADE20K dataset in $16^2$ non-retinal glimpse task. Our solution outperforms competitive solutions when only a small amount of the image is available.

*et al.*, 2021]. Note, that the authors of both baseline methods present their results as mean pixel accuracy (mPA - averaged over classes), while in fact they report pixel accuracy (PA - averaged over all pixels). We confirmed this finding with the authors directly. The qualitative results presented in Figure 7 further validate the superior performance.

### 5.4 Glimpse Analysis

**Number of glimpses.** The relationship between model performance in the reconstruction task and the number of allowed glimpses is illustrated in Figure 8. Our model consistently outperforms all convolutional baselines. In the case of MAE-based SimGlim [Jha *et al.*, 2023], our solution performs better until around 20% of patches are visited. This can be attributed to the fact that our solution optimizes the general reconstruction task, whereas the latter focuses on specific types of visited data when there is enough visual information already acquired.

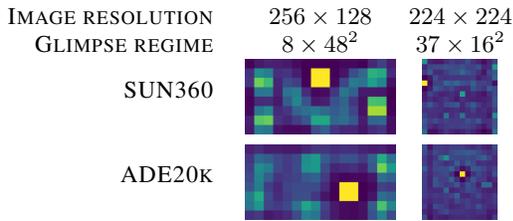

Figure 9: **Average glimpse image:** Mean glimpse map, averaged over all glimpse positions in all test images for reconstruction task on SUN360 and ADE20K. We can see that the first glimpse is consistently selected in the same place. Dataset and training regimes determine its best starting position.

| SELECTION | GLIMPSE REGIME | SUN360 | MS COCO |
|---|---|---|---|
| RANDOM | $8 \times 32^2$ | 32.8 | 38.3 |
| CHECKER | $8 \times 32^2$ | 32.0 | 32.6 |
| ATTENTION | $8 \times 32^2$ | **29.8** | **32.5** |
| RANDOM | $8 \times 16^2$ | 39.8 | 47.3 |
| CHECKER | $8 \times 16^2$ | 39.3 | 44.1 |
| ATTENTION | $8 \times 16^2$ | **37.9** | **43.2** |

Table 4: **AME-based glimpse selection versus naive sampling - reconstruction:** Comparison of glimpse selection strategies for reconstruction task on SUN360 and MS COCO databases. The metric used is a root mean square error (RMSE; lower is better). We used regular (non-retinal) glimpses for this ablation. Our method performs better than random position sampling and random sampling from a checkerboard-like pattern.

**Average glimpse selection.** The average glimpse location masks are shown in Figure 9. For the $8 \times 48^2$ glimpse setup, it can be observed that the glimpse selector prefers locations in the image corners and in the center. It rarely selects locations on the left and right edge for the SUN360 dataset, which may be because the dataset consists of $360°$ panoramic images. Selection masks for the $37 \times 16^2$ setup are much more evenly distributed, with only the first glimpse location being distinctive.

## 5.5 Glimpse Selection Strategies

Our attention-based glimpse selection process outperforms random glimpse selection and a fixed checkerboard-like glimpse pattern, as demonstrated in Table 4 and 5. This supports our claim that our selection strategy is superior to naive strategies and contradicts the discussions presented in [Jha *et al.*, 2023].

**Level of attention map.** In Figure 10, we demonstrate the impact of the choice of decoder layer as the attention map source on the model performance. The results show that the further in the decoder the attention maps are sourced from, the better the overall performance.

## 6 Conclusions

This paper presents a new approach to active visual exploration that simplifies the existing methods by leveraging internal model uncertainty in glimpse selection process. By utilizing the entropy of the transformer's attention maps, we elim-inate the need for any sampling decision module or separate network head focused on decision strategy. Furthermore, our approach does not require separate training for the main task and the sampling decision strategy, reducing the complexity of the overall system. Our approach is based on the masked auto-encoder architecture, which allows for the efficient processing of partial observations. The results demonstrate that the attention values trained inherently from the transformer-based decoder can be used to generate decision-making priors that improve the performance of scene understanding tasks. An exhaustive empirical comparison against the competing architectures shows that our solution performs better when limited data is available.

| SELECTION | GLIMPSE REGIME | TRAIN ALL | HEAD-ONLY |
|---|---|---|---|
| RANDOM | $8 \times 32^2$ | 72.4 | 66.2 |
| CHECKER | $8 \times 32^2$ | 70.1 | 64.0 |
| ATTENTION | $8 \times 32^2$ | **73.4** | **66.9** |
| RANDOM | $8 \times 16^2$ | 61.9 | 53.7 |
| CHECKER | $8 \times 16^2$ | 57.8 | **56.2** |
| ATTENTION | $8 \times 16^2$ | **63.4** | 55.9 |

Table 5: **AME-based glimpse selection versus naive sampling - classification:** Comparison of glimpse selection strategies in classification task for SUN360 database. The metric used is accuracy (higher is better). We used regular (non-retinal) glimpses for this ablation and tested full-model and head-only training strategies. Our method performs better than random position sampling and random sampling from a checkerboard-like pattern.

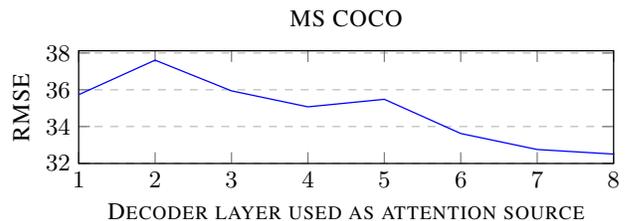

Figure 10: **RMSE results versus attention map source:** Comparison of attention-based glimpse selection results in regards to decoder layer used as attention map source. The figure shows that the best results can be achieved using the last layer of the decoder. The dataset used is MS COCO. The metric is a root mean square error (RMSE; lower is better). The image resolution is $256 \times 128$ in an $8 \times 32^2$ non-retinal glimpse regime.

## Acknowledgments

This research was funded by National Science Centre, Poland (grants no 2020/39/B/ST6/01511, 2022/45/B/ST6/02817, and 2021/41/B/ST6/01370), Foundation for Polish Science (grant no POIR.04.04.00-00-14DE/18-00 carried out within the Team-Net program co-financed by the European Union under the European Regional Development Fund). We gratefully acknowledge Polish high-performance computing infrastructure PLGrid (HPC Centers: ACK Cyfronet AGH) for providing computer facilities and support within computational grant no. PLG/2022/015753. The research was sup-ported by a grant from the Faculty of Mathematics and Computer Science under the Strategic Programme Excellence Initiative at Jagiellonian University.

## References

[Aloimonos *et al.*, 1988] John Aloimonos, Isaac Weiss, and Amit Bandyopadhyay. Active vision. *International journal of computer vision*, 1(4):333–356, 1988.

[Chai, 2019] Yuning Chai. Patchwork: A patch-wise attention network for efficient object detection and segmentation in video streams. *CoRR*, abs/1904.01784, 2019.

[Deng *et al.*, 2009] Jia Deng, Wei Dong, Richard Socher, Li-Jia Li, Kai Li, and Li Fei-Fei. Imagenet: A large-scale hierarchical image database. In *2009 IEEE conference on computer vision and pattern recognition*, pages 248–255. Ieee, 2009.

[Dosovitskiy *et al.*, 2020] Alexey Dosovitskiy, Lucas Beyer, Alexander Kolesnikov, Dirk Weissenborn, Xiaohua Zhai, Thomas Unterthiner, Mostafa Dehghani, Matthias Minderer, Georg Heigold, Sylvain Gelly, et al. An image is worth 16x16 words: Transformers for image recognition at scale. In *International Conference on Learning Representations*, 2020.

[Girdhar *et al.*, 2022] Rohit Girdhar, Mannat Singh, Nikhila Ravi, Laurens van der Maaten, Armand Joulin, and Ishan Misra. Omnivore: A Single Model for Many Visual Modalities. In *Proceedings of the IEEE/CVF International Conference on Computer Vision*, 2022.

[He *et al.*, 2022] Kaiming He, Xinlei Chen, Saining Xie, Yanghao Li, Piotr Dollár, and Ross Girshick. Masked autoencoders are scalable vision learners. In *Proceedings of the IEEE/CVF Conference on Computer Vision and Pattern Recognition*, pages 16000–16009, 2022.

[Jayaraman and Grauman, 2017] Dinesh Jayaraman and Kristen Grauman. Learning to look around. *CoRR*, abs/1709.00507, 2017.

[Jha *et al.*, 2023] Abhishek Jha, Soroush Seifi, and Tinne Tuytelaars. Simglim: Simplifying glimpse based active visual reconstruction. In *Proceedings of the IEEE/CVF Winter Conference on Applications of Computer Vision (WACV)*, pages 269–278, January 2023.

[Li *et al.*, 2022] Wenbo Li, Zhe Lin, Kun Zhou, Lu Qi, Yi Wang, and Jiaya Jia. Mat: Mask-aware transformer for large hole image inpainting. In *Proceedings of the IEEE/CVF Conference on Computer Vision and Pattern Recognition*, 2022.

[Lin *et al.*, 2014] Tsung-Yi Lin, Michael Maire, Serge Belongie, James Hays, Pietro Perona, Deva Ramanan, Piotr Dollár, and C Lawrence Zitnick. Microsoft coco: Common objects in context. In *European conference on computer vision*, pages 740–755. Springer, 2014.

[Loshchilov and Hutter, 2018] Ilya Loshchilov and Frank Hutter. Decoupled weight decay regularization. In *International Conference on Learning Representations*, 2018.

[Naseer *et al.*, 2021] Muzammal Naseer, Kanchana Ranasinghe, Salman H. Khan, Munawar Hayat, Fahad Shahbaz Khan, and Ming-Hsuan Yang. Intriguing properties of vision transformers. *CoRR*, abs/2105.10497, 2021.

[Pathak *et al.*, 2016] Deepak Pathak, Philipp Krahenbuhl, Jeff Donahue, Trevor Darrell, and Alexei A Efros. Context encoders: Feature learning by inpainting. In *Proceedings of the IEEE conference on computer vision and pattern recognition*, pages 2536–2544, 2016.

[Ramakrishnan and Grauman, 2018] Santhosh K. Ramakrishnan and Kristen Grauman. Sidekick policy learning for active visual exploration. *CoRR*, abs/1807.11010, 2018.

[Ramakrishnan *et al.*, 2021] Santhosh K. Ramakrishnan, Dinesh Jayaraman, and Kristen Grauman. An exploration of embodied visual exploration. *Int. J. Comput. Vis.*, 129:1616–1649, 2021.

[Rangrej and Clark, 2021] Samrudhdhi B. Rangrej and James J. Clark. Visual attention in imaginative agents. *CoRR*, abs/2104.00177, 2021.

[Rangrej *et al.*, 2022] Samrudhdhi B Rangrej, Chetan L Srinidhi, and James J Clark. Consistency driven sequential transformers attention model for partially observable scenes. In *Proceedings of the IEEE/CVF Conference on Computer Vision and Pattern Recognition*, pages 2518–2527, 2022.

[Sandini and Metta, 2003] Giulio Sandini and Giorgio Metta. Retina-like sensors: motivations, technology and applications. In *Sensors and sensing in biology and engineering*, pages 251–262. Springer, 2003.

[Seifi and Tuytelaars, 2019] Soroush Seifi and Tinne Tuytelaars. Where to look next: Unsupervised active visual exploration on 360° input. *CoRR*, abs/1909.10304, 2019.

[Seifi and Tuytelaars, 2020] Soroush Seifi and Tinne Tuytelaars. Attend and segment: Attention guided active semantic segmentation. *CoRR*, abs/2007.11548, 2020.

[Seifi *et al.*, 2021] Soroush Seifi, Abhishek Jha, and Tinne Tuytelaars. Glimpse-attend-and-explore: Self-attention for active visual exploration. In *Proceedings of the IEEE/CVF International Conference on Computer Vision*, pages 16137–16146, 2021.

[Śmieja *et al.*, 2018] Marek Śmieja, Łukasz Struski, Jacek Tabor, Bartosz Zieliński, and Przemysław Spurek. Processing of missing data by neural networks. *Advances in neural information processing systems*, 31, 2018.

[Song *et al.*, 2015] Shuran Song, Samuel P Lichtenberg, and Jianxiong Xiao. Sun rgb-d: A rgb-d scene understanding benchmark suite. In *Proceedings of the IEEE conference on computer vision and pattern recognition*, pages 567–576, 2015.

[Vaswani *et al.*, 2017] Ashish Vaswani, Noam Shazeer, Niki Parmar, Jakob Uszkoreit, Llion Jones, Aidan N Gomez, Łukasz Kaiser, and Illia Polosukhin. Attention is all you need. *Advances in neural information processing systems*, 30, 2017.

[Wenzel *et al.*, 2021] Patrick Wenzel, Rui Wang, Nan Yang, Qing Cheng, Qadeer Khan, Lukas von Stumberg, Niclas Zeller, and Daniel Cremers. 4seasons: A cross-season dataset for multi-weather slam in autonomous driving. In Zeynep Akata, Andreas Geiger, and Torsten Sattler, editors, *Pattern Recognition*, pages 404–417, Cham, 2021. Springer International Publishing.

[Zhou *et al.*, 2017] Bolei Zhou, Hang Zhao, Xavier Puig, Sanja Fidler, Adela Barriuso, and Antonio Torralba. Scene parsing through ade20k dataset. In *Proceedings of the IEEE conference on computer vision and pattern recognition*, pages 633–641, 2017.